\documentclass[journal]{IEEEtran}
\usepackage{cite}

\usepackage[hyphens]{url}
\usepackage{algorithm}
\usepackage[noend]{algpseudocode}

\ifCLASSINFOpdf
   \usepackage[pdftex]{graphicx}
\else
\fi
\usepackage{amsmath,amssymb,amsfonts}
\usepackage{xfrac}

\ifCLASSOPTIONcompsoc
 \usepackage[caption=false,font=normalsize,labelfont=sf,textfont=sf]{subfig}
\else
 \usepackage[caption=false,font=footnotesize]{subfig}
\fi
 \usepackage{dblfloatfix}
\usepackage{cleveref}

\begin{document}
%



\title{Interactive Constrained MAP-Elites: \\ \Large{Analysis and Evaluation of the Expressiveness of the Feature Dimensions}}

%
%
%


\author{Alberto~Alvarez,
        Steve~Dahlskog,
        Jose~Font,
        and~Julian~Togelius,~\IEEEmembership{Member,~IEEE}
\thanks{A. Alvarez, S. Dahlskog, and J. Font are with the Department of Computer Science and Media Technology (DVMT), Malmö University, Malmö, Sweden (e-mail: alberto.alvarez@mau.se; steve.dahlskog@mau.se; jose.font@mau.se).}
\thanks{J. Togelius is with the Department of Computer Science and Engineering, New York University, New York, NY 11201 USA (e-mail: julian@togelius.com).}
}

%
%

\markboth{IEEE Transaction on Games,~Vol.~XX, No.~X, Month~XXXX}%
{Alvarez \MakeLowercase{\textit{et al.}}: Interactive Constrained MAP-Elites}
%



\maketitle

\begin{abstract}
We propose the Interactive Constrained MAP-Elites, a quality-diversity solution for game content generation, implemented as a new feature of the Evolutionary Dungeon Designer (EDD): a mixed-initiative co-creativity tool for designing dungeons. The feature uses the MAP-Elites algorithm, an illumination algorithm that segregates the population among several cells depending on their scores with respect to different behavioral dimensions. Users can flexibly and dynamically alternate between these dimensions anytime, thus guiding the evolutionary process in an intuitive way, and then incorporate suggestions produced by the algorithm in their room designs. At the same time, any modifications performed by the human user will feed back into MAP-Elites, closing a circular workflow of constant mutual inspiration. This paper presents the algorithm followed by an in-depth 
evaluation of the expressive range of all possible dimension combinations in several scenarios, 
and discusses their influence in the fitness landscape and in the overall performance of the 
procedural content generation in EDD.

\end{abstract}


%
\IEEEpeerreviewmaketitle

\section{Introduction}


Procedural Content Generation (PCG) refers to the generation of game content with none or limited human input~\cite{Yannakakis2018}, where game content could be anything from rules and narrative, to levels, items, and music. While PCG has been a factor in game development since trailblazing games like~\emph{Rogue}~\cite{michael_toy_1980} and \emph{Elite}~\cite{braben_elite_1984}, it has only been a popular academic research topic for little more than a decade. Search-based PCG designates the use of a global search algorithm, such as an evolutionary algorithm to search content space~\cite{Togelius2011}.

Part of PCG's appeal is the promise to produce game art and content faster and at a lower cost, as well as enabling innovative content creation processes such as player-adaptive games~\cite{shaker2012evolving, hastings_evolving_2009, dormansUnexplored2017}, data-driven content generation~\cite{Khalifa2018, Green2018}, and mixed-initiative co-creativity~\cite{Liapis2016}. Mixed-initiative co-creativity (MI-CC), a concept introduced by Yannakakis et al.~\cite{yannakakis2014micc}, refers to the approach of using a creation process through which a computer and a human user provide 
and inspire each other in the form of iterative reciprocal stimuli. Examples of MI-CC systems are \textit{Pitako}~\cite{machado2019pitako}, \textit{Ropossum}~\cite{shaker2013ropossum}, \textit{Tanagra}~\cite{smith_tanagra:_2011}, \textit{CICERO}~\cite{Machado2017}, and \textit{Sentient Sketchbook}~\cite{liapis_generating_2013}. 

MI-CC aligns with the principles of lateral thinking and creative emotive reasoning: the processes of solving seemingly unsolvable problems or tackling non-trivial tasks through an indirect, non-linear, creative approach~\cite{Liapis2016}. Additionally, MI-CC provides insight on the affordances and constraints of the human process for creating and designing games~\cite{Yannakakis2018}.

A key mechanism in MI-CC approaches is to present suggestions to users
, and these suggestions must be of high quality but also be sufficiently diverse. So-called quality-diversity algorithms~\cite{Pugh2016} are very well suited for this, as they find solutions that have high quality according to some measure but are also diverse according to other measures~\cite{gravina2019procedural}. The Multi-dimensional Archive of Phenotypic Elites (MAP-Elites)~\cite{Mouret2015} is a suitable algorithm for this kind of problem. Khalifa et al.~\cite{Khalifa2018} presented constrained MAP-Elites, a combination MAP-Elites with the feasible-infeasible concept from the FI2Pop genetic algorithm~\cite{Kimbrough2008}, and applied this to procedurally generating levels for bullet hell games. 
Another recent implementation of MAP-Elites has been used to produce small sections of Super Mario Bros levels called \textit{scenes}, addressing specific game mechanics~\cite{Khalifa2019-intentionalCompLevel}.

The Evolutionary Dungeon Designer (EDD) is a MI-CC tool for generating dungeons for adventure games using a FI2Pop evolutionary approach~\cite{Alvarez2018, Alvarez2018a, Baldwin2017, Baldwin2017a}. This 
research was presented in~\cite{alvarez2019empowering}, 
introducing \emph{Interactive Constrained MAP-Elites} (IC MAP-Elites), a combination of Constrained MAP-Elites with interactive evolution, in the shape of 
a continuous evolutionary process that takes advantage of MAP-Elites' multidimensional discretization of the search space into cells. 
In \cite{alvarez2019empowering}, we analyzed the effects of using quality-diversity in procedurally generating dungeons, as well as the effects of continuous evolution and dimension customization.

This paper contributes with 
a thorough evaluation of the expressive range of EDD's MAP-Elites through all possible dimension combinations in several scenarios, as well as by extending the feature dimensions to 
two new dimensions, \emph{Inner similarity}, which adds a similarity measure independent of the aesthetics, and \emph{Leniency}, which strives to measure the subjective challenge score of a level. Following recent research on how to evaluate procedural content generators and, in particular, quality-diversity approaches~\cite{Cook2019:ParameterBasedEvaluation,Cook2016-SecondPaperDanesh,Gravina2019-blendingNotionsDiversity}, we present the results from new experiments 
with the objective to evaluate the expressive range of all dimensions in pairs, as well as to analyze how the generated and unique solutions relate to all the dimensions included in the search space, and assess IC MAP-Elites feasibility and adaptability to create dungeons and adventure levels.

\section{Previous Work}

\subsection{Map-Elites for illuminating search spaces}

Quality-diversity algorithms are algorithms which search a 
solution space, not just for the single best solution, but for a set of diverse solutions which are high performing. MAP-Elites maintains of map of good solutions~\cite{Mouret2015} and is a well-known quality-diversity algorithm
. The map is divided into a number of cells according to one or more feature dimensions. In each cell, a single solution is kept. At every update, an offspring is generated based on one or more existing solutions. That offspring     is then assigned to a cell based on its feature dimensions, which might or might not be the same as the cell(s) its parent(s) occupy. 
If the new offspring has a higher fitness than the existing solution in that cell, it replaces the previous item in the cell. This process results in a map of solutions where each cell contains the best found solution for those particular feature dimensions.

\subsection{Evaluation of Procedural Content Generators}
Shaker, et al.~\cite{shaker_procedural_2016}, argues that ultimately the evaluation of content generators is to verify that they fulfil their design goals. In order to be able to understand or modify a generator it is important to visualize its content space. However, it is seldom enough to look at a single individual piece of content, but rather it is vital to examine the frequency the different features appear, or the amount of variety the features demonstrate. Previous attempts of doing this are termed expressivity measures~\cite{Smith:2010:Expressive-range,Summerville2018-ExpresiveRange} and have, for instance, explored difficulty measures~\cite{Horn2014-comparativePCG}. Other approaches incorporate tool assisted parameter exploration due to its effect on the content space~\cite{Cook2016-danesh,Cook2019:ParameterBasedEvaluation}.


\subsection{Evolving Dungeons as a Whole, Room by Room}


EDD is a MI-CC tool that allows a human designer to create a 2D dungeon and the rooms it is composed of
. The designer is able to manually edit both the dungeon by placing and removing rooms, and the individual rooms by  editing the tiles 
that each room consists of. EDD's underlying evolutionary algorithm provides procedurally generated suggestions, and is driven through the use of game design micro- and meso- patterns
. A detailed description of all EDD's features, including the use of game design patterns, can be found in~\cite{Baldwin2017a, Baldwin2017, Alvarez2018, Alvarez2018a}, where \cite{Alvarez2018,Baldwin2017} analyze and discuss the mixed-initiative system in EDD and \cite{Baldwin2017a,Alvarez2018a} focus on the procedural content generator.


\begin{figure}[t]
\centerline{\includegraphics[width=0.47\textwidth]{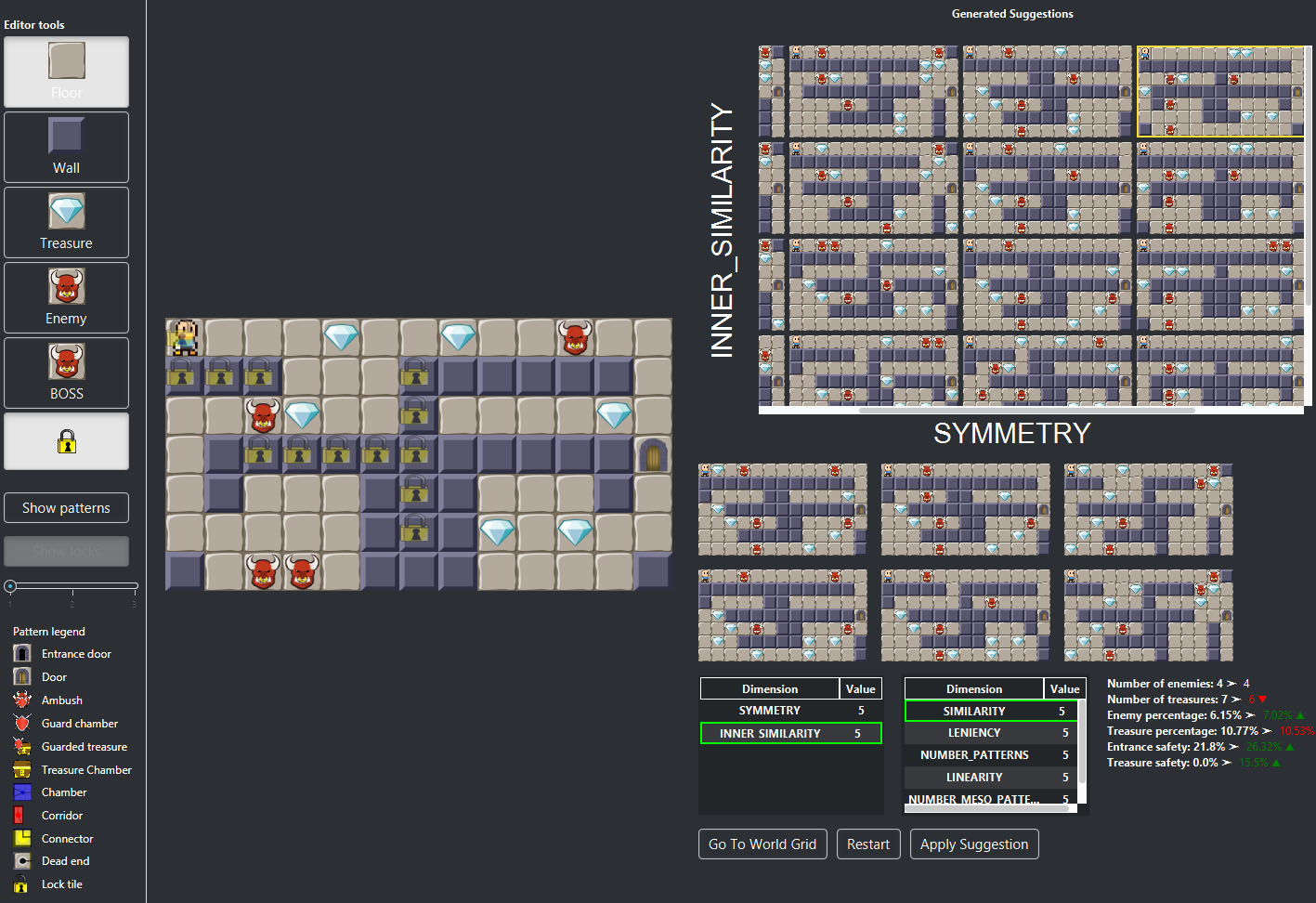}}
\caption{The room editor screen in EDD. The left pane contains all the options for manually editing the room displayed at the center-left of the screen. The right section displays the procedurally generated suggestions.}
\label{figs:roomscreen}
\end{figure}

In this section we present the latest version of EDD\footnote{Available for download at \url{https://github.com/mau-games/eddy}}, which includes significant improvements based on the outcomes from the qualitative analysis discussed in~\cite{Alvarez2018}. The dungeon is now represented as a graph of interconnected rooms of any given size between $3\times3$ and $20\times20$ tiles. The smallest allowed dungeon is composed by two rooms with one connection to each other. Rooms and connections can now be added and removed at any time. Connections are marked with door tiles.

The designer marks one room as the \textit{initial room} for feasibility calculation: a dungeon is feasible when there is at least one path between the \textit{initial room} and any other passable tile in the dungeon. Rooms and doors that are unreachable from the \textit{initial room} are highlighted in red, so that they can be easily identified by the designer.



The starting screen in EDD is the dungeon editor screen
. Every new room is empty (composed solely of floor tiles) when created and is placed detached from the dungeon graph. After manually connecting the room to the dungeon with at least one connection, the designer has the option to populate the room using the room editor screen (\Cref{figs:roomscreen}). This screen can be reached by either double-clicking on the room or by clicking on the "Start with our suggestions" button, which will present six procedurally generated suggestions to the designer to start editing from. 



\Cref{figs:roomscreen} shows the room editor screen displaying a sample room with the dimensions $7\times13$ tiles. The left pane lists all the available options for manually editing the room by brush painting with one of the available tile types (floor, wall, treasure, or enemy) and one of the two brush sizes (single tile and five-tile cross shape). Control-clicking allows the designer to bucket paint all adjacent tiles of the same type. Brush painting with the lock button on preserves selected tiles in all the procedurally generated suggestions. A detailed description of all the options in this pane is included in~\cite{Alvarez2018, Alvarez2018a}.

The right side of the screen displays the procedurally generated suggestions, by means of the Interactive Constrained MAP-Elites (IC MAP-Elites) genetic algorithm (see \Cref{section:3}). 
When the designer accesses the room editor screen, IC MAP-Elites starts and continuously populates the suggestions pane with elites. The evolutionary process is fed with the manually edited room (i.e. target room), so that every change in the room affects the generated suggestions. By clicking on "Apply Suggestion", the manually edited room is replaced by the selected suggestion, thus affecting the upcoming procedural suggestions. "Restart" restarts evolution, and "Go To World Grid" takes the user back to the dungeon editor screen.
\section{Interactive Constrained MAP-Elites} \label{section:3}

The overarching goal of MI-CC is to collaborate with the user to produce content, either to optimize (i.e., exploit) their current design towards some goal or to foster (i.e., explore) their creativity by surprising them with diverse proposals. By implementing MAP-Elites~\cite{Mouret2015} and continuous evolution into EDD, our algorithm can (1) account for the multiple dimensions that a user can be interested in, (2) explore multiple areas of the search space and produce a diverse amount of high-quality suggestions to the user, and (3) still evaluate how interesting and useful the tile distribution is within a specific room. Henceforth, we name the presented approach~\textbf{Interactive Constrained MAP-Elites} (IC MAP-Elites).

EDD uses a single-objective fitness function (shown in~\Cref{eq:fitness_func}) with a FI2Pop genetic algorithm, where fitness is a weighted sum divided equally between (1) the inventorial aspect of the rooms and (2) the spatial distribution of the design. $f_{inventorial}$ is the evaluation of the aggregated and normalized quality of treasures, enemies, and doors (inventorial patterns). $f_{spatial}$ refers to the quality and distribution of chambers, i.e. open areas in the room and corridors (both categorized as spatial patterns), and the meso-patterns that are created within chambers and their quality. Quality refers to the positioning, safety, composition, and relation between patterns. The fitness adapts to the user's current design, automatically informing target ratios and distributions to be used as targets. In-depth evaluation of EDD's fitness function, as well as discussion and explanation of the quality of each inventorial, spatial, and meso pattern, can be found in~\cite{Alvarez2018a,Baldwin2017,Baldwin2017a}.



\begin{equation} 
\label{eq:fitness_func}
f_{fitness}(r) = \frac{1}{2}f_{inventorial}(r) \,+ \, \frac{1}{2}f_{spatial}(r)
\end{equation}

Furthermore, IC MAP-Elites adds interactive and continuous evolution to the Constrained MAP-Elites presented by Khalifa et al.~\cite{Khalifa2018}. This is done through an adaptive fitness function (based on the designer's design) that adapts the content generation, and by enabling the designer to flexibly change the feature dimensions and the granularity of the cells. It also adapts the usability of MAP-Elites to generate dungeon and adventure levels in an MI-CC system, which gives more control to the designers over non-intuitive parameters and aspects of MAP-Elites, while providing a richer set of high-performing and diverse suggestions.

\subsection{Illuminating Dungeon Populations with MAP-Elites}

MAP-Elites explores the search space more vastly by separating interesting feature dimensions, that affect different aspects of the room, such as playability or visual aesthetics, from the fitness function, to categorize rooms into cells. 

In this subsection, we firstly present all the current feature dimensions identified and implemented in EDD, secondly, we explain the transition from fixed evolution to continuous evolution, and lastly, we introduce and outline the IC MAP-Elites algorithm.

\subsubsection{Dimensions}\label{sec:dimensions}
\begin{figure}[h]
\centering
     \subfloat[Basic room\label{figs:targetRoomsBas}]{%
       \includegraphics[width=0.15\textwidth]{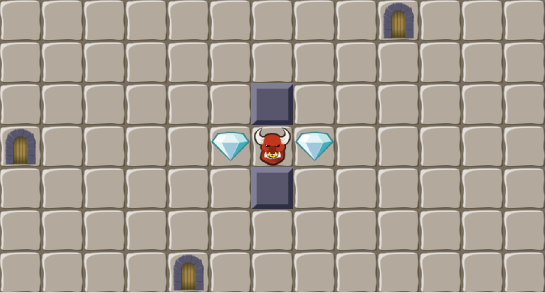}
     }
     \subfloat[Complex room\label{figs:targetRoomsComp}]{%
       \includegraphics[width=0.15\textwidth]{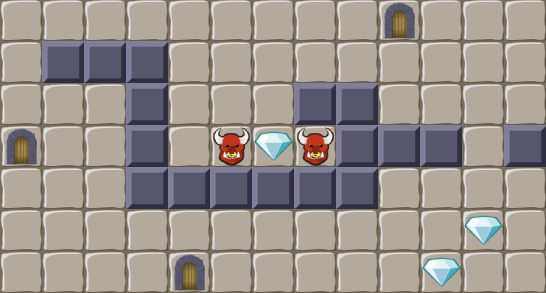}
     }\hfill
     \subfloat[Example Elites generated using IC MAP-Elites with (b) as target. \label{figs:dimensions-example}]{%
       \includegraphics[width=0.45\textwidth]{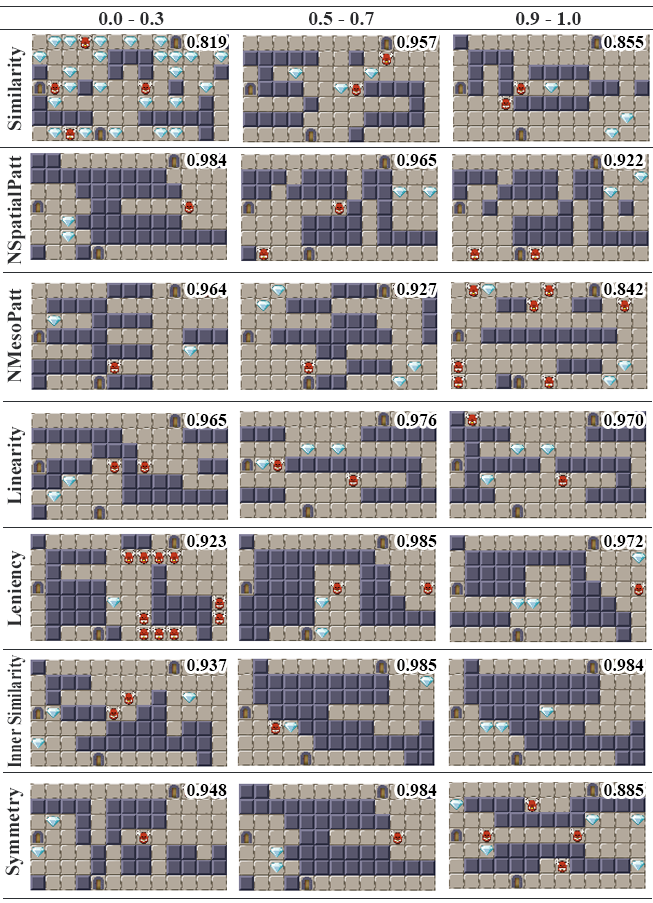}
     }\hfill

\caption{(a) and (b) represent target rooms used in the experiments. Each row in (c) represents an independent run of the algorithm using the dimension specified to the left. Each column splits the dimension score into three intervals. Each cell displays (top-right) the fitness of the optimal individual in its related interval.
}
\end{figure}

Dimensions in MAP-Elites are identified as those aspects of the individuals that can be calculated in the behavioral space, and that are independent of the fitness calculation. These are crucial in MAP-Elites, as they represent the discretized dimensions where individuals will be retained as the space is explored. 
EDD offers the designer the possibility to choose among the following dimensions, two at a time (examples of rooms generated using the dimensions are shown in~\Cref{figs:dimensions-example}):

\textbf{Symmetry (Sym) and Similarity (Sim).} We choose Symmetry as a consideration of the aesthetic aspects of the edited room since symmetric structures tend to be more visually pleasing for the user and relate to fairness distribution and human-made structures~\cite{marinho2016-symmetry,Alvarez2018a,Liapis2012-adaptiveVisual}. Similarity is used to present the user variations of their design but still preserving their aesthetical edits. Symmetry is assessed using only impassable tiles (i.e., walls) and evaluated along the X and Y axes and diagonals. The highest value is normalized by the total amount of walls and used as the symmetry score (\Cref{eq:Symmetry}). Similarity is calculated by comparing tile by tile with the target room (\Cref{eq:similarity}). In-depth descriptions and evaluations can be found in~\cite{Alvarez2018a}, where both dimensions were used as aesthetic fitness evaluations.



\begin{equation} \label{eq:Symmetry}
D_{sym} = \frac{highestSymmetricValue} {totalWalls}
\end{equation}

\begin{equation} \label{eq:similarity}
D_{sim} = \frac{totalTiles - notSimilarTiles} {totalTiles}
\end{equation}

\textbf{Number of Meso-patterns (NMP).} The number of meso-patterns correlates to the type and amount of encounters the designer wants the user to have in the room in a more structured manner. The considered patterns are the treasure room (tr), guard rooms (gr), and ambushes (amb). Meso-patterns associate utility to a set of tiles in the room, for instance, a long chamber filled with enemies and treasures could be divided into 2 chambers, the first one with enemies and the second one with treasures so the risk-reward encounter is more understandable for the player. Since we already analyze the rooms for all possible patterns, the number of meso-patterns is simply $\#MesoPat=tr, gr, amb \in AllPatterns$. \Cref{eq:meso-pat-eq} presents the dimensional value, and since the used meso-patterns can only exist in a chamber, we normalize by the maximum amount of chambers in a room, which are of a minimum size of $3\times3$, and results in $Max_{chambers}=\left\lfloor Cols/3 \right\rfloor \cdot \left\lfloor Rows/3 \right\rfloor$.



\begin{equation} \label{eq:meso-pat-eq}
D_{NMP} = \min \left\{ \dfrac{\#MesoPat}{Max_{chambers}}, 1.0 \right\}
\end{equation}

\textbf{Number of Spatial-patterns (NSP).} By spatial-patterns we mean chambers (c), corridors (cor), connectors (con), and nothing (n). We identify the number of spatial-pattern relates to how individual tiles group (or not) together to form spatial structures in the room. The higher the amount of spatial-patterns the lesser tiles will be group together in favor of more individualism. For instance, a room with one spatial-pattern can be one with no walls and just an open chamber, while a room with a higher number of spatial-patterns would subdivide the space with walls, using tiles for more specific patterns. \Cref{eq:spatial-pat-eq} presents how we calculate the value for such a dimension. The number of spatial patterns is simply $\#SpatialPat=c, n, cor, con \in AllPatterns$, we then normalize it by the largest side of the room and multiply it by a constant value, determined as $K=4.0$ through a process of experimentation.




\begin{equation} 
\label{eq:spatial-pat-eq}
D_{NSP} = \min\left\{\frac{\#SpatialPat}{\max\left\{{Cols, Rows}\right\} \cdot \textit{K}}, 1.0\right\}
\end{equation}

\textbf{Linearity (Lin).} Linearity represents the number of paths that exist between the doors in the room. This relates to the type of gameplay the designer would like the room to have by the distributions of walls among the room. Having high linearity in a room does not need to only be by having a narrow corridor between doors but could also be generated by having all doors in the same open space (i.e. the user would not need to traverse other areas) or by simply disconnecting all paths between doors. \Cref{eq:Linearity-eq} shows the linearity calculation. Due to the use of patterns, we calculate the paths between doors as the number of paths that exist from a spatial-pattern containing a door to another. Finally, this is normalized by the number of spatial patterns in combination with the number of doors and their possible neighbors.

\begin{equation} \label{eq:Linearity-eq}
D_{lin} = 1 \text{--} \frac{AllPathsBetweenDoors} {\#spatialPat + \#NeighborsPerDoor}
\end{equation}


\textbf{Inner Similarity (IS).} Inner similarity compares the target room to one generated, considering only the distribution and ratios of micro-patterns in both rooms rather than any aesthetic criteria. Specifically, we look into the density ($den$) and sparsity ($spa$) of enemies ($en$), treasures ($tre$), and walls ($wal$) in the target room ($R_{tg}$) in comparison with the generated rooms ($R_{gen}$). To calculate the density and sparsity of each micro-pattern, we first clustered all micro-patterns of the same kind based on the distance within the room (i.e. $distance=1$). We then use these clusters to calculate the density (\Cref{eq:dens}) using as density threshold $\theta=4$ for treasures and enemies, and $\theta=6$ for walls, and calculate the sparsity (\Cref{eq:spars}). Finally, we calculate the difference between each distribution in $R_{tg}$ and $R_{gen}$, linearly combining all the values into the IS measure as in \Cref{eq:innersimDim}.

\begin{equation} \label{eq:dens}
den(x)= \dfrac{\sum_{i=1}^{\left | clusts \right |}\min \left\{1.0, \dfrac{\left | clust_{i} \right |}{\theta_{x}}\right\}}{\left | clusts \right |}
\end{equation}

\begin{equation} \label{eq:spars}
spa(x)= \dfrac{\sum_{i=1}^{\left | clusts \right |}\sum_{j=1, j\neq i}^{\left | clusts \right |}\dfrac{Dist(clust_{i}, clust_{j})}{\left | room \right |}}{\left | clusts \right | \cdot (\left | clusts \right | - 1)}
\end{equation}

\begin{equation} \label{eq:innersimDim}
\begin{split}
D_{IS} = \sum_{i=1}^{micropats}\left | den(R_{gen}) - den(R_{tg}) \right | + \\ \left | spa(R_{gen}) - spa(R_{tg}) \right |
\end{split}
\end{equation}

\textbf{Leniency (Len).} Leniency calculates how challenging a room is at any given point. It is based on the amount of enemies and treasures that are in a room, their density and sparsity calculated as in eq. (\ref{eq:dens}) and (\ref{eq:spars}), respectively, and how safe the doors are (i.e. entry/exit points) calculated as in ~\cite{Baldwin2017}. We base our calculation in the idea that rooms are less lenient the more enemies they contain as well as how they are distributed, counterbalanced by the number of treasures as they reward players. We calculate the dimension value for each room as shown in (\Cref{eq:lendim}), which uses a combination of precomputed non lenient (\Cref{eq:nonlenvals}) and lenient (\Cref{eq:lenvals}) values.

\begin{equation} \label{eq:nonlenvals}
\begin{split}
nonLenientValues = w_{0} \cdot \log_{10}(\left | en \right | \cdot spa(en)) + \\ w_{1} \cdot \log_{10}(\left | en \right | \cdot den(en)) + w_{2} \cdot (1.0 - door_{safety})
\end{split}
\end{equation}

\begin{equation} \label{eq:lenvals}
\begin{split}
LenientValues = \sfrac{1}{2} \log_{10}(\left | tre \right | \cdot spa(tre)) + \\ \sfrac{1}{2} \log_{10}(\left | tre \right | \cdot den(tre))
\end{split}
\end{equation}

\begin{equation} \label{eq:lendim}
D_{len} = 1.0 - (nonLenientValues - (\sfrac{1}{2} \cdot LenientValues))
\end{equation}



\subsubsection{Continuous and Interactive Evolution}



Since EDD already uses a FI2Pop, we took the Constrained MAP-Elites, presented by Khalifa et al.~\cite{Khalifa2018}, as a starting point. The illuminating capabilities of MAP-Elites explore the search space with the constraints aspects of FI2Pop. This approach manages two different populations, a feasible and an infeasible one, within each cell. Individuals move across cells when their dimension values change, or between the feasible and infeasible population according to their fulfillment of the feasibility constraint.

As discussed by Takagi, interactive evolution is a way to improve the capabilities of Evolutionary Algorithms (EA) by having humans in-the-loop to subjectively evaluate individuals. This hybrid approach has proven to reach better and more adaptive results but at the expenses of user's fatigue and user's understanding of the EA and the given problem~\cite{Takagi2001-InteractiveEvo}. However, in EDD and the IC MAP-Elites, the user does not directly evaluate individuals; instead, IC MAP-Elites adapts the fitness and search based on different interactions the designer has with the algorithm. 

The designer can interact with the crossover step by locking tiles~\cite{Alvarez2018a}, and with the dimensions and cells by changing the dimensions and granularity for the MAP-Elites, enabling IC MAP-Elites to focus on different regions of the generative space. Furthermore, IC MAP-Elites constantly updates the target room and configuration with the most recent version of the designer's design. Once the suggestions are broadcasted, that room is incorporated without changes to the population of individuals in the corresponding cell.

This adaptability feature and different designer's indirect interactions with IC MAP-Elites, enables the implementation of continuous and interactive evolution, as well as allowing the designer to focus solely in the design of the room, while IC MAP-Elites adapts to the new designs.


\algnewcommand\algorithmicforeach{\textbf{for each}}
\algdef{S}[FOR]{ForEach}[1]{\algorithmicforeach\ #1\ \algorithmicdo}

\algblockdefx{MRepeat}{EndRepeat}{\textbf{repeat}}{}
\algnotext{EndRepeat}

\begin{algorithm}
\footnotesize
\caption{Interactive Constrained MAP-Elites}\label{alg:IC-MAPE}
\begin{algorithmic}[1]
\Procedure{IC MAP-Elites($\protect[\{d_1,v_1\},...,\{d_n,v_n\}]$)}{}
\State $target \gets curEditRoom$ \Comment{Always in background}
\State createCells$(\protect[\{d_1,v_1\},...,\{d_n,v_n\}])$
\For{$i \gets 1$ to $PopSize$} 
     \State add mutate$(target)$ to $population$
\EndFor
\State CheckAndAssignToCell$(population)$ 
\While {true} \Comment{start continouous evo}
    \For{$generation \gets 1$ to $publishGen$}
        \If {$\textit{dimensionsChanged}$}
            \State $previousPop \gets cells_{pop}$
            \State createCells$(newDimensions)$
            \State checkAndAssignToCell$(previousPop)$ 
        \EndIf
        \MRepeat{ \text{[for feasible \& infeasible pop.]}}
            \For{$i \gets 1$ to $ParentIteration$}
                \State $curCell \gets \text{rndCell}(cells)$
                \State add tournament$(curCell)$ to $parent$
            \EndFor
            \State $offspring \gets  \text{crossover}(Parent)$
            \State checkAndAssignToCell$(offspring)$
        \EndRepeat
        \State sortAndTrim$(cells)$
    \EndFor
    \State broadcastElites() \Comment{render elites}
    \State $pop' \gets cells_{population}$
    \State add mutate$(cells_{pop})$ to $pop'$
    \State add $target$ to $pop'$
    \State checkAndAssignToCell $(pop')$
    \State sortAndTrim$(cells)$
\EndWhile
\EndProcedure
\Procedure{createCells(dimensions)}{}
    \ForEach{$dim \in dimensions $}
        \State add newCell$(dim_d, dim_v)$ to $cells$
    \EndFor
\EndProcedure
\Procedure{$\protect \text{check\&AssignToCell}(curPopulation)$}{}
    \ForEach{$individual \in curPopulation $}
        \State $individual_f \gets evaluate(individual)$ 
        \State $individual_d \gets dim(individual)$
        \State add $individual$ to $cell_{pop}(individual_d)$
    \EndFor
\EndProcedure
\end{algorithmic}
\end{algorithm}


\subsubsection{Algorithm}

IC-MAP-Elites is depicted in~\Cref{alg:IC-MAPE}. Cells are first created based on the dimensions selected by the user and proceed to initialize the population based on the user's design, evaluate it and assign each individual to the corresponding cell. Before starting each generation, we check if the dimensions have changed, and if so, recreate the cells and populate them with the previous individuals, and proceed through the evolutionary strategies. We first select uniformly random which cell to choose parents from, and then we select 5 parents through tournament-selection. Offspring are produced through a two-point uniform crossover operation with a 30\% chance of mutation. Offspring are placed in the correct cell and population after calculating their fitness and dimension's information. Finally, cells eliminate the low-performing individuals that over-cap their maximum capacity. Since interbreeding is not allowed, and can only happen indirectly (i.e. the offspring changing population and then used for breeding in consequent generations), the strategies are repeated for each of the populations.

IC MAP-Elites runs for $n$ generations, and once it reaches the specified limit, it broadcasts the found elites. In order to foster the exploration, we first mutate all the individuals from all the populations and cells (while retaining the previous population), and add them into the same pool together with the current edited room without changes. Finally, we evaluate and assign all the individuals to the correct cells, and cells that are over maximum capacity eliminates low-performing individuals. This procedure is repeated until the user decides to stop the algorithm.
\section{Performance Evaluation across Dimensions}

\begin{figure}[h]
\centerline{\includegraphics[width=0.48\textwidth]{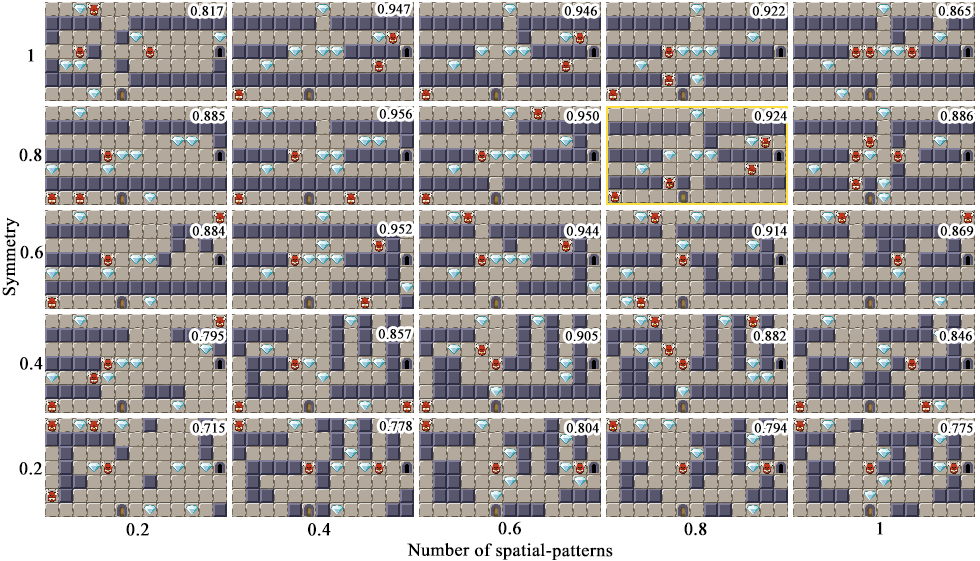}}
\caption{Rooms at generation $2090$ targeting Number of spatial-patterns (X) and Symmetry (Y). Each cell displays (top-right) the fitness of the optimal individual in its related feasible population. }
\label{figs:patt_sym}
\end{figure}

First, we ran a set of experiments to test the results from the IC MAP-Elites using all possible combinations of the available dimensions using two dimensions at a time. All experiments were run using $13\times7$ rooms, the same room size as in \emph{The Binding of Isaac}~\cite{mcmillen_binding_2011}, a representative example of a dungeon-based adventure game. In each experiment, the initial population was set to $1000$ mutated individuals distributed in feasible and infeasible populations in all cells which were set to a maximum capacity of $25$ individuals each. IC MAP-Elites ran continuously, and every $100$ generations rendered the elites of each cell. At each generation, it selected $5$ parents per population among uniformly random chosen cells. Offspring were always produced through a two-point crossover and had a 30\% chance of being mutated, which would randomly alter one tile in the level.

\Cref{figs:patt_sym} shows a grid containing the best found suggestions at generation $2090$, while aiming for number of spatial-patterns at the X-axis and symmetry at the Y-axis with a granularity of $5$. Each cell displays the optimal individual of the feasible population under a given pair of dimension values. The fitness score is displayed on the cells' top-right corner.

The fitness evaluation in IC MAP-Elites is quite lightweight in terms of computational cost, which enables the grid of suggestions to be completed in a matter of seconds. This is of principal importance for successfully implementing continuous evolution, so that the influence of each manual change in the edited rooms is reflected in the suggestions almost instantly. 

Results in \Cref{figs:patt_sym} are representative of the good quality diversity solutions produced by EDD. The average fitness across elites is $0.872$, and the highest fitness is $0.956$ (cell $[0.4,0.8]$). No two rooms are the same. As intended, high levels of symmetry are displayed in the upper rows, gradually decreasing towards the bottom row. Similarly, rooms in the leftmost column contain lower amounts of spatial patterns, increasing towards the rightmost column. Lower amounts of spatial patterns translate into more open rooms with almost no corridors and one or two large adjacent chambers (as in cell $[0.2, 0.2]$), as opposed to highly pattern filled rooms that comprise intricate pathways converging at one or two small chambers (cell $[1, 0.2]$). 

Fitness values show that some dimension combinations are harder to optimize than others, so that the whole grid depicts a gradient landscape of the compatibility between each pair of dimensions. The bottom-left corner shows difficulties producing symmetric rooms with low amounts of spatial patterns, as opposed to rooms with many corridors (upper-right corner), which seem to favor the generation of symmetrical structures.

Additional similar experiments can be found in~\cite{alvarez2019empowering}, where we combined other pairs of dimensions and analyzed and discussed the correlation and limitations found in them. However, in the following section, we present an extension of those evaluations through a more exhaustive and in-depth assessment of IC MAP-Elites.




\section{Expressive Range Analysis}

\begin{table}[]
\caption{Comparison of the avg. explored space and avg. fitness of the generated individuals in different evaluations.}
\label{tab:evaluationTable}
\begin{tabular}{|l|lll|}
\hline
Evaluation & \multicolumn{1}{c|}{$\Diamond$} & \multicolumn{1}{c|}{$\bigtriangleup$} & \multicolumn{1}{c|}{$\bigcirc$} \\ \hline
\textbf{IC MAP-Elites}: Avg. dimensions in pairs & 0.89 & 36.46\% & 52.4\% \\ \hline
\textbf{IC MAP-Elites}: All Dimensions & 0.78 & 51.7\% & N/A      \\ \hline
\textbf{Objective-based EA}& 0.92 & 22.48\% & N/A      \\ \hline

\multicolumn{4}{l}{$\Diamond$ Avg. Fitness \ \
$\bigtriangleup$ Avg. coverage across all dimensions} \\ 
\multicolumn{4}{l}{$\bigcirc$ Avg. coverage in respective dimension pair}
\end{tabular}%
\end{table}

\begin{figure*}[h!]
\centerline{\includegraphics[width=15cm]{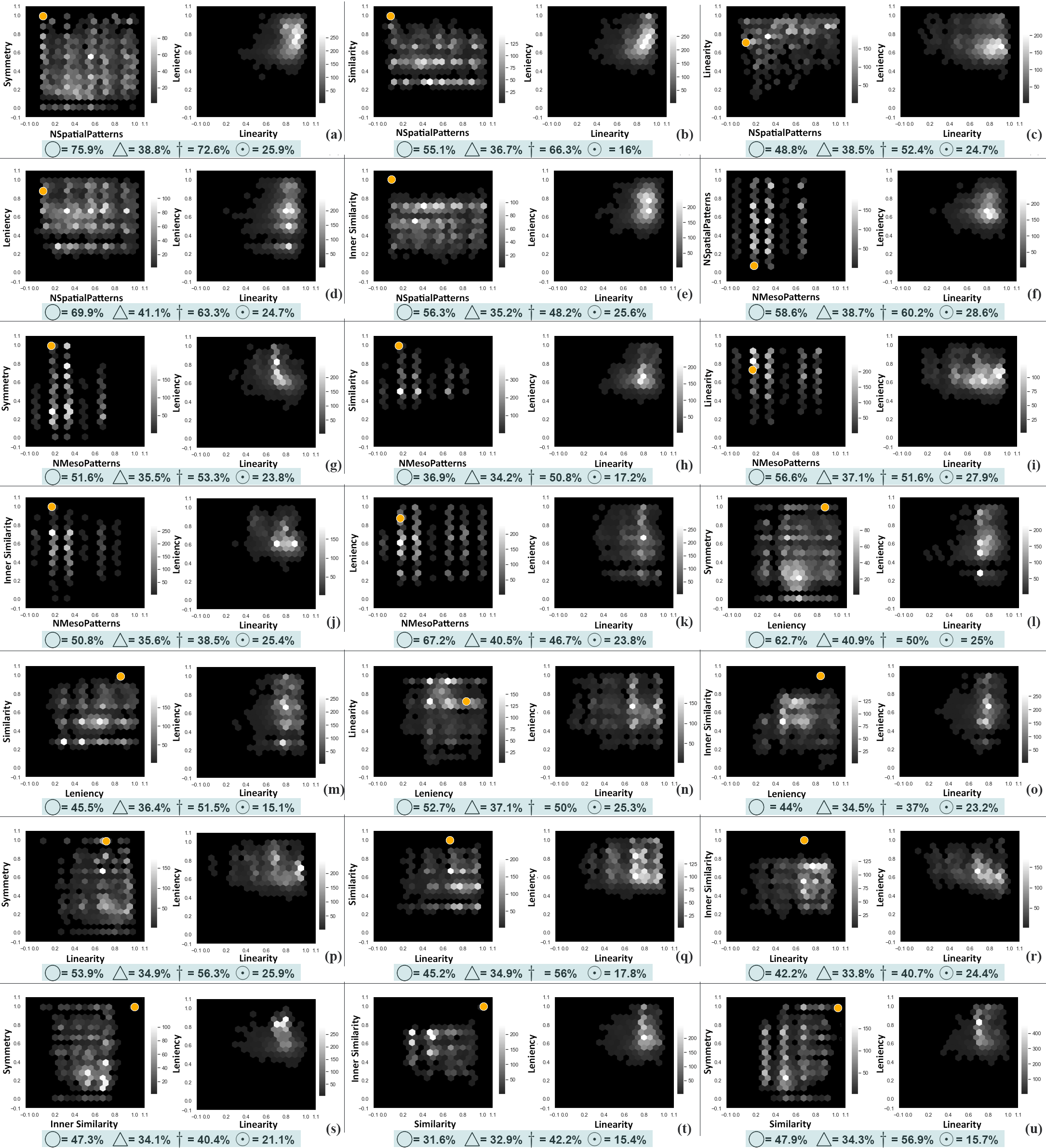}}
\caption{Expressive range of the 21 possible combinations (from a to u) of dimensions picked in pairs. Each subfigure is composed of two plots: (left) an evaluation based on a given pair of dimensions; (right) the same pair of dimensions but evaluated in terms of Linearity and Leniency scores. Each hexagon relates to a dimensional score, whereas a lighter hue indicates a higher number of unique individuals generated under that particular score. All runs used~\Cref{figs:targetRoomsBas} as the target room, and its dimensional score is highlighted with an orange mark. Under each subfigure, we present data particular to the specific tested pair of dimensions. $\bigcirc$, $\dagger$, and $\odot$ represent the coverage percentage in the respective pair of dimensions when running IC MAP-Elites with the subfigure's pair of dimensions ($\bigcirc$), IC MAP-Elites with all dimensions in the search ($\dagger$), and when using objective-based EA ($\odot$). $\bigtriangleup$ represents the avg. coverage percentage across all dimensions when running IC MAP-Elites with the subfigure's pair of dimensions.
}
\label{figs:full-expressive}
\end{figure*}


\begin{figure*}[h!]
\centerline{\includegraphics[width=\textwidth]{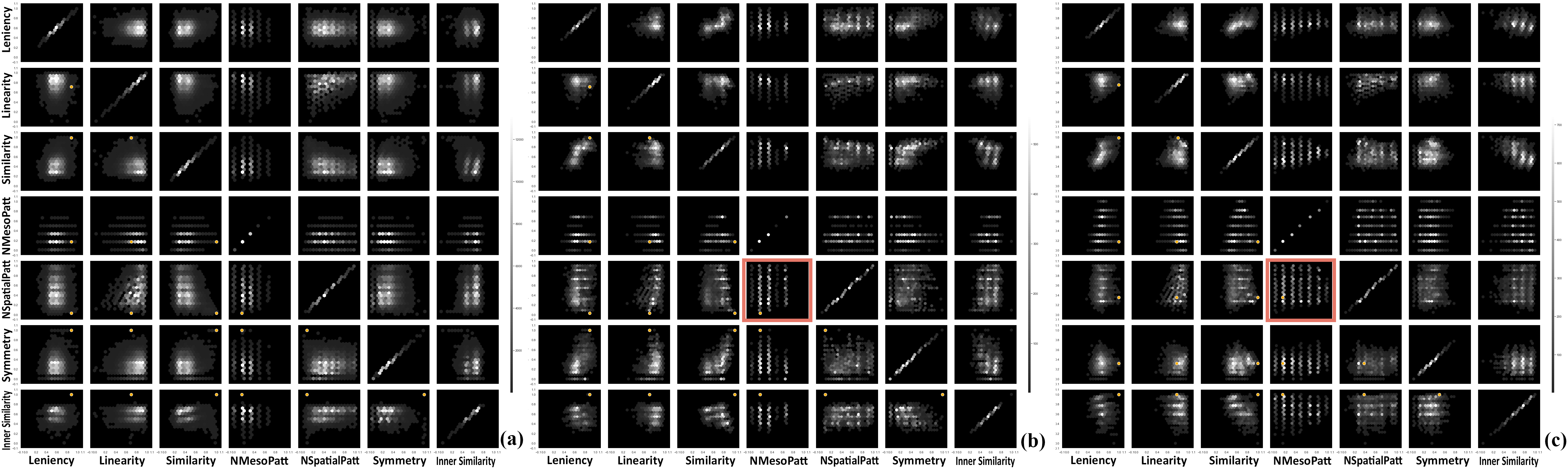}}
\caption{In-detail run showing how all dimensions relate to each other in alternative scenarios. In (a) IC MAP-Elites ran for 2000 generations using the same target room as in \Cref{figs:full-expressive} (\Cref{figs:targetRoomsBas}), but using all the dimensions in the search, which results in 78125 cells. (b) Shows how all dimensions were explored when using~\Cref{figs:full-expressive}f (NMP-NSP) as pair of dimension. In (c) we used the same dimensions as in (b) but we changed the target room to~\Cref{figs:targetRoomsComp}. The dimensional score of the respective target room in each subfigure (\Cref{figs:targetRoomsBas} for a and b, and \Cref{figs:targetRoomsComp} for c) is highlighted with an orange mark. In b and c, the dimensions used for the experiment are highlighted with a red border.}
\label{figs:all-dimensions-earun}
\end{figure*}

\begin{figure}[h]
\centerline{\includegraphics[width=8.5cm]{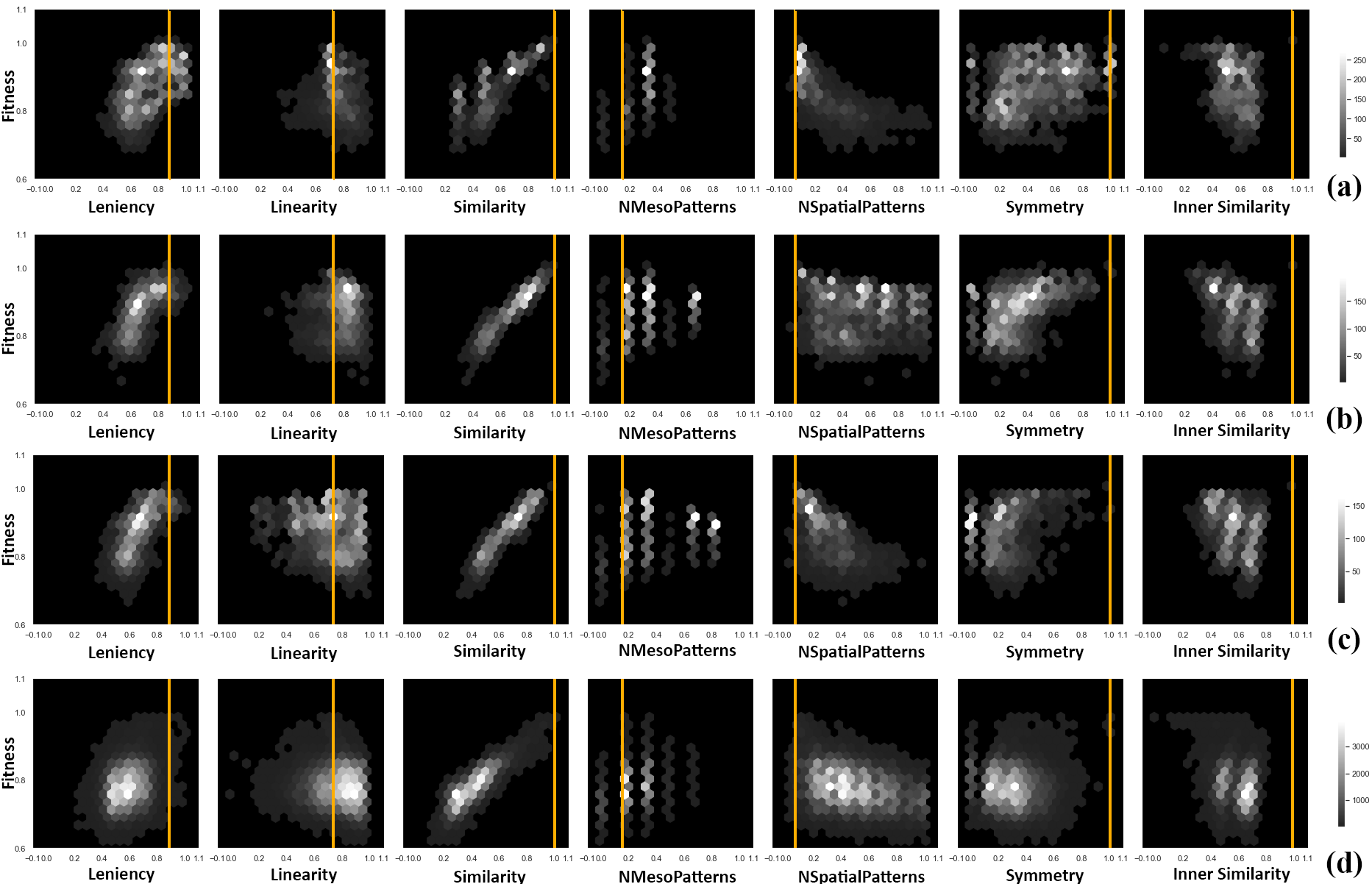}}
\caption{Relation between dimension score and fitness score. (a), (b), and (c) show results using, respectively, Similarity and Symmetry, NMP and NSP, and NMP and Linearity. (d) was run using all the 7 dimensions. All runs used~\Cref{figs:targetRoomsBas} as target room, and it's dimensions score are denoted with an orange line across the different plots.}
\label{figs:dimensions-related-fitness}
\end{figure}

We ran a second set of experiments analyzing the expressive range \cite{Smith:2010:Expressive-range} of the IC MAP-Elites using the 21 possible combinations of dimension pairs. We followed the same setup as presented in the previous section, and every $100$ generations, we collected the unique generated individuals' data, which was aggregated in all the expressive ranges and individually plotted in~\Cref{figs:avgfitness}. Through this, between $150$ to $2001$ individuals were produced every $100$ generations. By exploring the expressive range and conducting a comparative analysis among the various dimension combinations, we intend (1) to assess IC MAP-Elites, its exploration and exploitation capabilities, and the feature dimensions, (2) to analyze how different dimension combinations affect the search for QD content, and (3) to identify bias in the search space.

Moreover, in~\Cref{tab:evaluationTable} we present a comparison of the avg. diversity and avg. quality of the generated individuals between different approaches to support our evaluation. The approaches consist of IC MAP-Elites picking dimensions in pairs (avg. of the aggregated results), IC MAP-Elites using all dimensions at the same time in the search, and the objective-based EA that was used in EDD previous to IC MAP-Elites. When using a pair of dimensions in IC MAP-Elites, the search explores in avg. $52.4$\% in the respective pair of dimensions (individual results are presented in~\Cref{figs:full-expressive}) and the individuals generated have an avg. fitness of $0.89$. In comparison, when using all dimensions rather than only pairs, the search explores $51.7$\% across all dimensions, $15$\% more than when using pairs, but with a drop in the avg. fitness of the population ($0.78$), which is discussed further in~\Cref{sec:ERAFitness} and shown in~\Cref{figs:dimensions-related-fitness}. As expected and concluded before by Mouret and Clune~\cite{Mouret2015}, objective-based EA generated individuals with an avg. high fitness ($0.92$), but with low diversity ($22.48$\%). This means that the search had a narrower focus, and the generated levels did not differ much from each other. Objective-based ran for $5000$ generations, but it stagnated and stopped generating novel levels after $13$ generations i.e., stopped exploring, generating a total of $1669$ levels. Conversely, IC MAP-Elites kept generating novel levels until stopped, but considerably less after $1000$ generations.

Figure \ref{figs:full-expressive} shows the expressive range of the IC MAP-Elites with each letter referring to a unique pair of dimensions tested, and the subfigure divided into two different plots. In the left plot, we evaluate the setup based on the used pair of dimensions with each hexagon placed in relation to their dimensions' score. The hue of each of them is connected to the number of unique suggestions generated. Likewise, in the right plot, we evaluated the setup based on its linearity and leniency score, which is used to compare the setups' expressiveness. All the setups were run for $5000$ generations, using the same target room, which is shown as an orange marker (\Cref{figs:targetRoomsBas}).

In~\Cref{figs:full-expressive}, it is shown that with IC MAP-Elites, it is explored a substantial area of the generative space (denoted in the figure with $\bigcirc$ under each subfigure) rather than just exploiting the area around the target room, depicted as an orange marker. On average, in all the independent runs in their respective dimensions, the search explores around $52$\% of the space, filling at least half of the map with high-performing elites averaging $0.89$ (as shown in~\Cref{tab:evaluationTable}). It can also be observed that the dense areas of the search space (i.e., where the algorithm exploited the most) are distant from the target room's scores, and most of them are sparse throughout the search. This indicates that using IC MAP-Elites and a pair of dimensions at a time helps the distribution of the search while exploiting promising areas, and the search is less likely to be biased towards creating levels similar to the target.

Nevertheless, when using both leniency and linearity to compare the performance of the dimensions' pairs, it is shown that they are underexplored and within the same range (0.4-1.0) when not using them as dimensions. This points towards the search having difficulties getting out of other dimensions' local optima, especially since the densest search area is within the target room (this is denoted with $\bigtriangleup$ under each subfigure).

\subsection{Alternative Scenarios}

In \Cref{figs:all-dimensions-earun}, we examine how the algorithm would vary its dimensions' exploration and exploitation in two different scenarios. (a) Using the same target room as in all the cases in \Cref{figs:full-expressive} but using all the possible dimensions in the search space (i.e. 7), and (c) using NMP and NSP (see Section~\ref{sec:dimensions}) as dimensions in the search but changing the target room to~\Cref{figs:targetRoomsComp}. To draw a better comparison, we added (b), which shows how all dimensions were explored when using~\Cref{figs:full-expressive}f (NMP-NSP) as pair of dimension.



When using all the dimensions (a), IC MAP-Elites can explore a substantial area of the search space in each of the dimensions ($51.7$\%), which in total is 15\% more searched space than when using only a pair of dimensions on average. This is expected since all the dimensions are now acting as archives. However, as it is noted in~\Cref{figs:full-expressive} under each subfigure, the actual explored space in the respective pairs ($\bigcirc$) is most of the time greater than what using all dimensions explore in the respective pair ($\dagger$).

It can also be observed that when using NMP and NSP as dimensions~\Cref{figs:all-dimensions-earun}b and c, regardless of the target room, IC MAP-Elites manages to still search a good amount of space (in avg. $38.7$\% and $43$\% among all dimensions, respectively). We suspect that this is because the range between low and high scores in the NMP or NSP dimensions produces very different rooms, as it can be seen in \Cref{figs:dimensions-example} in their respective rows. 


Moreover, when comparing (a) and (b), it is noticeable that while (a) explores a greater area than (b) (in general, $15.24$\% more), 
it seems to be recurrently generating the same type of individuals (i.e. depicted with the hue of the hexagon) while in (b), the dense areas for most of the dimensions are sparser, especially when matching the pair of dimensions used for evaluation. 
 Finally, it should be noted that while these three plots are comparable in their diversity search, they differ in the number of elites they store during the search, with an archive of 78125 cells for (a), and 25 cells for (b) and (c).


\subsection{Fitness Evaluation} \label{sec:ERAFitness}

Figure~\ref{figs:dimensions-related-fitness} shows the relation of the fitness with the explored individuals in each dimension in 4 independent runs. (a, b, c) Were runs using dimensions in pairs with the same data and dimensions as in \Cref{figs:full-expressive} (u), (f), and (h), respectively. (d) Was run using all the dimensions in the search space with the same data as in \Cref{figs:all-dimensions-earun} (a). There is an evident high correlation between Similarity scores and fitness across all the subfigures, which is expected since our fitness value is highly dependant on the target's ratios. In contrast, IS is not even close to match the fitness curve of Similarity rather there are high-performing individuals along the dimension, even when IS calculates similarity using ratios, densities, and sparsities of the target's micropatterns to calculate the score of individuals. 

Moreover, our experiment shows that when using specific dimensions (\Cref{figs:dimensions-related-fitness}a-c), we achieve a relatively better search (i.e. find more diverse and high-quality individuals) in those dimensions, while still being able to explore the rest of dimensions. For instance, when not using NSP as a feature dimension such as in (a) or (c), the NSP dimension is fully explored but generating no high-quality individuals, meanwhile, when using NSP as a feature dimension as in (b), the search can find individuals in the same range as in (a) or (c) but with a higher fitness. 
Similar results can be seen in the rest of the dimensions where the search uses specific dimensions, for instance, in (a) exploring diverse and higher quality individuals in Similarity, or in (c) in Linearity.


The most interesting result can be seen in (d), where we used all dimensions in the search. This allows for a vast search of diverse individuals in all dimensions, but at the same time it seems to exploit sub-optimal areas. On the other hand, when using only a pair of dimensions as in (a), (b) and (c), the search remained dense in high-quality individuals in all dimensions.


\begin{figure}[h]
\centerline{\includegraphics[width=7cm]{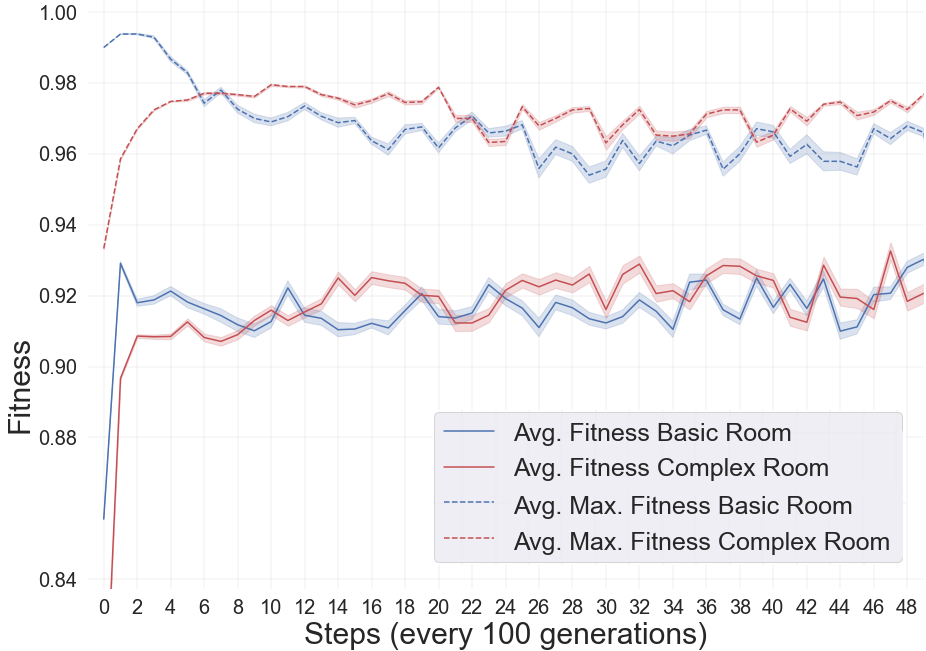}}
\caption{Fitness evolution per 100 generations throughout all the runs. 
Blue represents the fitness of the basic room (\Cref{figs:targetRoomsBas}), and Red represents the fitness of the complex room (\Cref{figs:targetRoomsComp}). 
}
\label{figs:avgfitness}
\end{figure}

Furthermore, figure~\ref{figs:avgfitness} shows the aggregated fitness over time of all dimension pairs (21). The avg. max fitness and the avg. total fitness are depicted as a dashed line and a continuous line, respectively. Both are surrounded by a low opacity thicker line representing the confidence interval. The graph shows that across all pairs of dimensions, the avg. fitness of novel generated levels is very high regardless of the target room and the generation, with a minor fluctuation between $0.92$ and $0.94$. In addition, the graph furthers supports what is presented in \Cref{figs:dimensions-related-fitness} a, b, and c, with most of the individuals generated in high score regions of the space in relation to fitness, but also showing that this is stable throughout the generations. The stable fitness 

This stationary overall high fitness is expected from MAP-Elites as it aims at constantly generating high-performing diverse individuals. This diversity goal, combined with an adaptive fitness dependant on the target room, makes it harder to generate levels with maximum fitness.~\Cref{figs:avgfitness} clearly shows this; depending on the fitness, there are scores in dimensions or pairs of dimensions that are less compatible with the fitness evaluation. Yet, what has been shown thus far in the multiple expressive range analysis and especially in~\Cref{figs:dimensions-related-fitness}a-c, is that IC MAP-Elites is able to generate high-performing and diverse levels. 






\section{Conclusions and Future Work}





In this paper, we have done an in-depth evaluation of IC MAP-Elites by analyzing the expressive range of all the possible pairs of dimensions, their relation to other dimensions and the fitness. 
Our results indicate specific dimensions in level generation, such as when using NMP, NSP, \textsc{leniency}, or Symmetry, that foster greater exploration. The exploration is not only fostered in their respective dimensions (in avg. 54\%, 61\%, 57\%, 57\%, respectively, when used) but also in all the others due to the diverse individuals generated within the respective dimensions as shown in \Cref{figs:dimensions-example}. As observed on~\Cref{figs:dimensions-related-fitness}, aesthetic feature dimensions such as~\emph{Similarity} and~\emph{Symmetry} have an impact when used or not in the search. When not using~\emph{Symmetry}, the search does not explore high levels of symmetry, disregarding to some extent that aesthetic feature in favor of exploring the other feature dimensions. Moreover, \emph{Similarity} has a high correlation with fitness as observed in~\Cref{figs:dimensions-related-fitness} and depending on the designer's objective, this might affect positively or negatively since there will be a strong bias towards highly similar individuals. In contrast, \textsc{is} seems to be more robust in the fitness landscape and the exploration of other dimensions because it captures the properties of the target room rather than its aesthetics. 

Regarding the bias in the exploration when using IC MAP-Elites together with continuous and adaptive evolution, our results show that the generated content is highly diverse with dense areas along most of the searched space, which is shown as well in \Cref{tab:evaluationTable}. This means that due to the diversity pressure imposed by IC MAP-Elites, the search is unlikely to be biased towards creating content that is similar in the feature dimensions' scores of the target room. Yet IC MAP-Elites adapts to the target room and generates high-performing individuals along the rest of the space in the other dimensions, especially in those explicitly used in the search.

To further assess the algorithm, we ran an experiment using all possible dimensions (\Cref{figs:all-dimensions-earun}a) rather than specific pairs to observe the exploration and exploitation of the algorithm when not using specific pair of dimensions. As expected, it explores a substantial area of the search space in all dimensions (in avg. $51.7$\%) but surprisingly, the search results in the exploitation of sub-optimal individuals in all dimensions with an avg. fitness of $0.78$ as shown in \Cref{figs:dimensions-related-fitness}d. While using pair of dimensions results on $15.24$\% reduced exploration, the exploitation seems to be fairly spread among the explored space, visible in \Cref{figs:all-dimensions-earun}b and c, and the density related to fitness is focused on high-performing individuals~\Cref{figs:dimensions-related-fitness}a-c. This points towards that there are difficulties in (1) \textbf{fully exploring} the space when using a pair of dimensions even when the exploitation is distributed and focused on high-performing individuals, and in (2) \textbf{exploiting the promising areas} of the search when using a higher range of dimensions even when the space is vastly explored. 

In addition, based on our experiments, such difficulties are exacerbated since the exploration stagnates and keeps exploiting the same areas after \~{}1000 generations, yet finding novel individuals. This happens regardless of the number of dimensions, which dimensions, or the target room. Our findings point to challenges in the \emph{selection step} of IC MAP-Elites, which selects cells uniformly random. Exploring different methods for the selection of cells and individuals is a promising future step. For instance, Gravina et al. \cite{Gravina2019-blendingNotionsDiversity} explored how four divergent search algorithms guide cells' selection.
benefit MAP-Elites standard selection method.


Furthermore, preliminary experiments seem to indicate that some dimensions, which are more explorative (e.g., NMP, NSP, Leniency, Symmetry), are more robust to changes in the target room, exploring similar areas of the search space regardless of the target, which can be observed in \Cref{figs:all-dimensions-earun} (b) and (c). This points towards dimensions that would make the algorithm more robust to changes, reinforcing its adaptability feature. Further experiments are needed to analyze how different dimensions are better at adapting to continuous changes in the target room, which would also indicate better stability in the search. Along these lines, further evaluation is needed with human designers to assess and explore whether IC MAP-Elites is beneficial for the MI-CC workflow and interaction. 


Currently, IC MAP-Elites is only used in EDD on a per room basis. Future work should consider the whole dungeon for the evolution procedure and further develop for the generation of complete dungeons using holistic metrics to evaluate the dungeon.


In conclusion, based on our results, IC MAP-Elites generate more diverse levels while retaining high-quality among them, which would result in richer and more options and suggestions to designers. Our experiments show that which dimensions are used significantly impact the search space, fostering the search of high-quality individuals within the selected dimensions while not discouraging exploration in other dimensions. This means that by editing their levels and choosing which dimension IC MAP-Elites should focus on, the designers will be given more meaningful choices and interactions.

\section*{Acknowledgment}
The Evolutionary Dungeon Designer is part of the project \textit{The Evolutionary World Designer}, supported by The Crafoord Foundation.

\ifCLASSOPTIONcaptionsoff
  \newpage
\fi



\bibliographystyle{IEEEtran}
\bibliography{references.bib}
\end{document}